\documentclass[10pt,twocolumn,letterpaper]{article}

\usepackage{cvpr}
\usepackage{times}
\usepackage{epsfig}
\usepackage{graphicx}
\usepackage{amsmath}
\usepackage{amssymb}
\usepackage{url}
\usepackage{booktabs}
\usepackage{multirow}
\usepackage{subfigure}
\usepackage{lib}

\usepackage[pagebackref=true,breaklinks=true,letterpaper=true,colorlinks,bookmarks=false]{hyperref}

\cvprfinalcopy 

\def\cvprPaperID{****} 

\newcommand{\pool}[1]{$\textrm{pool}_{#1}$\xspace}
\newcommand{\conv}[1]{$\textrm{conv}_{#1}$\xspace}
\newcommand{\maxp}[1]{$\textrm{max}_{#1}$\xspace}
\newcommand{\fc}[1]{$\textrm{fc}_{#1}$\xspace}

\title{Deformable Part Models are Convolutional Neural Networks\\{\large Tech report}}

\author{Ross Girshick~~~~Forrest Iandola~~~~Trevor Darrell~~~~Jitendra Malik\\
UC Berkeley\\
{\tt\small \{rbg,forresti,trevor,malik\}@eecs.berkeley.edu}}

\begin{document}

\maketitle

\begin{abstract}
Deformable part models (DPMs) and convolutional neural networks (CNNs) are two widely used tools for visual recognition.
They are typically viewed as distinct approaches:
DPMs are graphical models (Markov random fields), while CNNs are ``black-box'' non-linear classifiers.
In this paper, we show that a DPM can be formulated as a CNN, thus providing a novel synthesis of the two ideas.
Our construction involves unrolling the DPM inference algorithm and mapping each step to an equivalent (and at times novel) CNN layer.
From this perspective, it becomes natural to replace the standard image features used in DPM with a learned feature extractor.
We call the resulting model DeepPyramid DPM and
experimentally validate it on PASCAL VOC.
DeepPyramid DPM significantly outperforms DPMs based on histograms of oriented gradients features (HOG) and slightly outperforms a comparable version of the recently introduced R-CNN detection system, while running an order of magnitude faster.
\end{abstract}

\section{Introduction}
\seclabel{intro}

Part-based representations are widely used for visual recognition tasks.
In particular, deformable part models (DPMs) \cite{lsvm-pami} have been especially useful for generic object category detection.
DPMs update pictorial structure models \cite{Felzenszwalb05,Fischler73} (which date back to the 1970s) with modern image features and machine learning algorithms.
Convolutional neural networks (CNNs) \cite{fukushima1980neocognitron,lecun-89e,rumelhart86} are another influential class of models for visual recognition.
CNNs also have a long history, and have come back into popular use in the last two years due to good performance on image classification \cite{decafICML,krizhevsky2012imagenet} and object detection \cite{girshick2014rcnn,overfeat} tasks.

These two models, DPMs and CNNs, are typically viewed as distinct approaches to visual recognition.
DPMs are graphical models (Markov random fields), while CNNs are ``black-box'' non-linear classifiers.
In this paper we describe how a DPM can be formulated as an equivalent CNN, providing a novel synthesis of these ideas.
This formulation (DPM-CNN) relies on a new CNN layer, \emph{distance transform pooling}, that generalizes max pooling.
Another innovation of our approach is that rather than using histograms of oriented gradients (HOG) features \cite{Dalal05}, we apply DPM-CNN to a feature pyramid that is computed by another CNN.
Since the end-to-end system is the function composition of two networks, it is equivalent to a single, unified CNN.
We call this end-to-end model \emph{DeepPyramid DPM}.



We also show that DeepPyramid DPM works well in practice.
In terms of object detection mean average precision, DeepPyramid DPM slightly outperforms a comparable version of the recently proposed R-CNN \cite{girshick2014rcnn} (specifically, R-CNN on the same \conv{5} features, without fine-tuning), while running about 20x faster. 
This experimental investigation also provides a greater understanding of the relative merits of region-based detection methods, such as R-CNN, and sliding-window methods like DPM.
We find that regions and sliding windows are complementary methods that will likely benefit each other if used in an ensemble.



HOG-based detectors are currently used in a wide range of models and applications, especially those where region-based methods are ill-suited (poselets \cite{BourdevMalikICCV09} being a prime example).
Our results show that sliding-window detectors on deep feature pyramids significantly outperform equivalent models on HOG.
Therefore, we believe that the model presented in this paper will be of great practical interest to the visual recognition community.
An open-source implementation will be made available, which will allow researchers to easily build on our work.

\section{DeepPyramid DPM}
\seclabel{sec_model}

In this section we describe the DeepPyramid DPM architecture.
DeepPyramid DPM is a convolutional neural network that takes as input an image pyramid and produces as output a pyramid of object detection scores.
Although the model is a single CNN, for pedagogical reasons we describe it in terms of two smaller networks whose function composition yields the full network.
A schematic diagram of the model is presented in \figref{model}.


\subsection{Feature pyramid front-end}

Objects appear at all scales in images.
A standard technique for coping with this fact is to run a detector at multiple scales using an image pyramid.
In the context of CNNs, this method dates back to (at least) early work on face detection in \cite{lecun94}, and has been used in contemporary work including OverFeat \cite{overfeat}, DetectorNet \cite{DetectorNet}, and DenseNet \cite{densenet}.
We follow this approach and use as our first CNN a network that maps an image pyramid to a feature pyramid.
We use a standard single-scale architecture (Krizhevsky \etal \cite{krizhevsky2012imagenet}) and tie the network weights across all scales.
Implementation details are given in \secref{impl}.

\begin{figure*}[t!]
\centering
\includegraphics[scale=0.6]{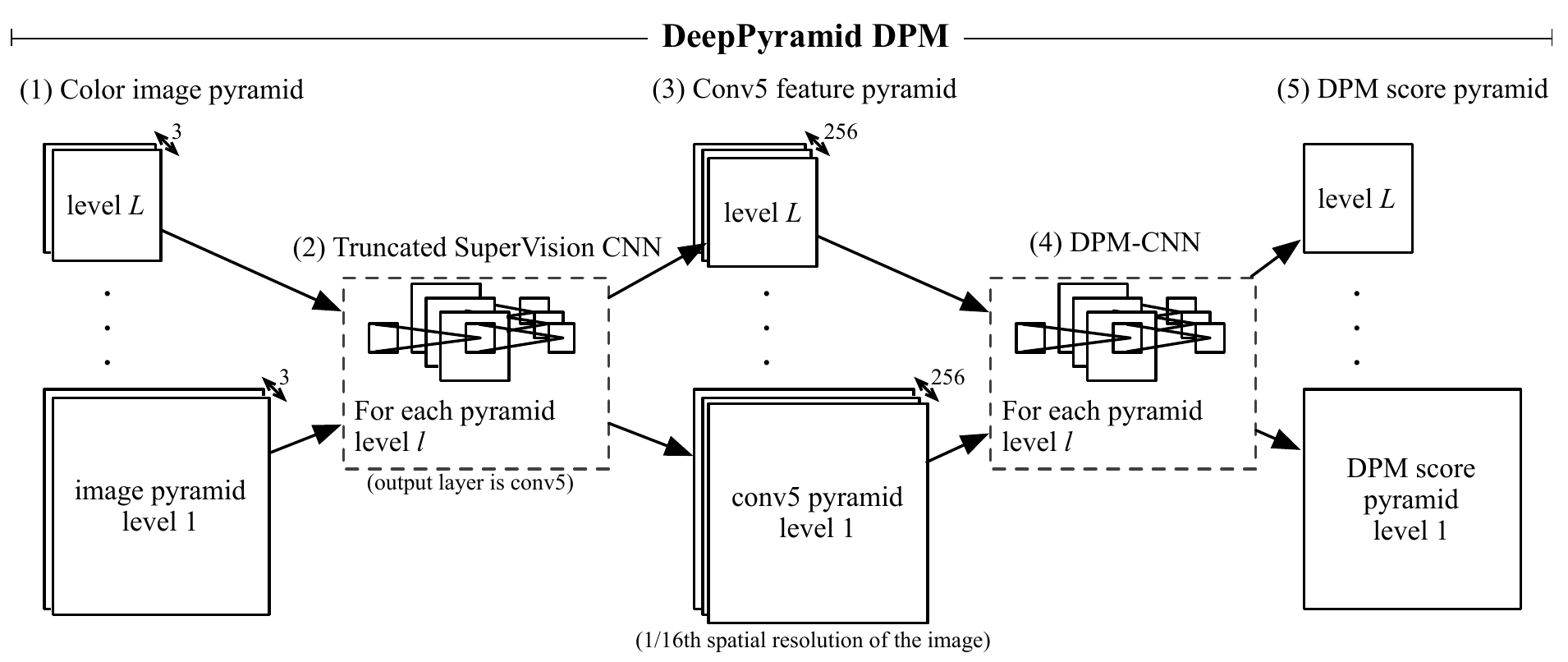}
\caption{\textbf{Schematic model overview.} 
(1) An image pyramid is built from a color input image. 
(2) Each pyramid level is forward propagated through a truncated SuperVision CNN \cite{krizhevsky2012imagenet} that ends at convolutional layer 5 (\conv{5}). 
(3) The result is a pyramid of \conv{5} feature maps, each at 1/16th the spatial resolution of its corresponding image pyramid level. 
(4) Each \conv{5} level is then input into a DPM-CNN, which produces a pyramid of DPM detection scores (5).
Since the whole system is simply the composition of two CNNs, it can be viewed as a single, unified CNN that takes a color image pyramid as input and outputs a DPM score pyramid.
}
\figlabel{model}
\end{figure*}

\subsection{DPM-CNN: Constructing an equivalent CNN from a DPM}

In the DPM formalism, an object class is modeled as a mixture of ``components'', each being responsible for modeling the appearance of an object sub-category (\eg, side views of cars, people doing handstands, bi-wing propeller planes).
Each component, in turn, uses a low-resolution global appearance model of the sub-type (called a ``root filter''), together with a small number (\eg, 8) of higher resolution ``part filters'' that capture the appearance of local regions of the sub-type.
At test time, a DPM is run as a sliding-window detector over a feature pyramid, which is traditionally built using HOG features (alternatives have recently been explored in \cite{lim2013sketch,HSC}). 
DPM assigns a score to each sliding-window location by optimizing a score function that trades off part deformation costs with image match scores.
A global maximum of the score function is computed efficiently at all locations by sharing computation between neighboring positions using a dynamic programming algorithm.
This algorithm is illustrated with all of the steps ``unrolled'' in Figure 4 of \cite{lsvm-pami}.
The key observation in this section is that for any given DPM, its unrolled detection algorithm yields a specific network architecture with a fixed depth.

In order to realize DPM as a convolutional network, we first introduce the idea of distance transform (DT) pooling, which generalizes the familiar max-pooling operation used in CNNs.
Given DT-pooling, a single-component DPM is constructed by composing part-filter convolution layers and DT-pooling layers with an ``object geometry'' layer that encodes the relative offsets of DPM parts.
We call this architecture DPM-CNN; its construction is explained in \figref{dpm-cnn-comp}.
To simplify the presentation, we consider the case where the DPM parts operate at the same resolution as the root filter.
Multi-resolution models can be implemented by taking two scales as input and inserting a subsampling layer after each DT-pooling layer.

\subsubsection{Distance transform pooling}

Here we show that distance transforms of sampled functions \cite{dt} generalize max pooling.

First, we define max pooling.
Consider a function $f{:}~\mathcal{G} \rightarrow \mathbb{R}$ defined on a regular grid $\mathcal{G}$.
The max pooling operation on $f$, with a window half-length of $k$, is also a function $M_f{:}~\mathcal{G} \rightarrow \mathbb{R}$ that is defined by $M_f(p) = \max_{\Delta p \in \{-k, \ldots, k\}} f(p+\Delta p)$.

Following \cite{dt}, the distance transform of $f$ is a function $D_f{:}~\mathcal{G} \rightarrow \mathbb{R}$ defined by $D_f(p) = \max_{q \in \mathcal{G}} \left(f(q) - d(p-q)\right)$. 
In the case of DPM, $d(r)$ is as a convex quadratic function $d(r) = ar^2 + br$, where $a > 0$ and $b$ are learnable parameters for each part.
Intuitively, these parameters define the shape of the distance transform's pooling region.

Max pooling can be equivalently expressed as $M_f(p) = \max_{q \in \mathcal{G}} \left(f(q) - d_{\textrm{max}}(p-q)\right)$,
where $d_{\textrm{max}}(r)$ is zero if $r \in \{-k, \ldots, k\}$ and $\infty$ otherwise.
Expressing max pooling as the maximization of a function subject to a distance penalty $d_{\textrm{max}}$ makes the connection between distance transforms and max pooling clear.
The distance transform generalizes max pooling and can introduce learnable parameters, as is the case in DPM.
Note that unlike max pooling, the distance transform of $f$ at $p$ is taken over the entire domain $\mathcal{G}$.
Therefore, rather than specifying a fixed pooling window a priori, the shape of the pooling region can be learned from the data.


In the construction of DPM-CNN, DT-pooling layers are inserted after each part filter convolution.

\begin{figure*}[t!]
\centering
\includegraphics[scale=0.6]{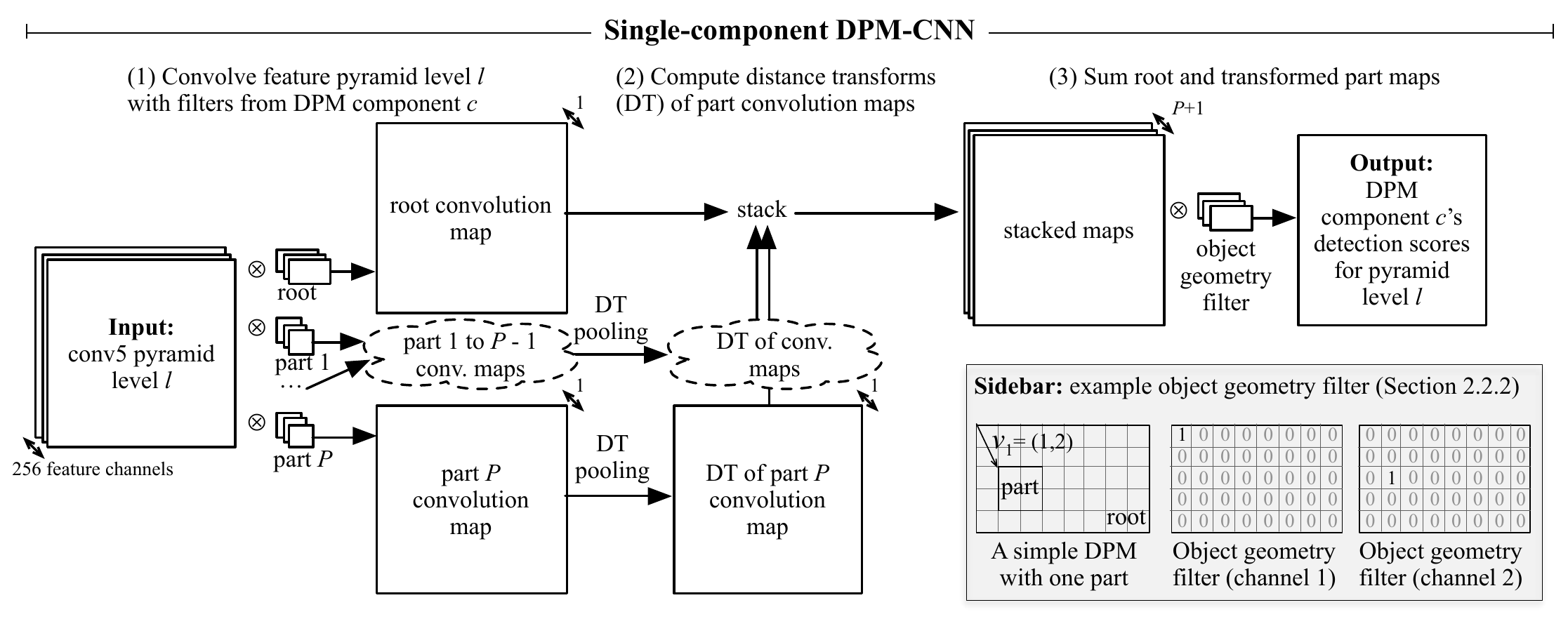}
\caption{\textbf{CNN equivalent to a single-component DPM.} 
DPM can be written as an equivalent CNN by unrolling the DPM detection algorithm into a network.
We present the construction for a single-component DPM-CNN here and then show how several of these CNNs can be composed into a multi-component DPM-CNN using a maxout layer (\figref{dpm-cnn-multicomp}).
A single-component DPM-CNN operates on a feature pyramid level.
(1) The pyramid level is convolved with the root filter and $P$ part filters, yielding $P+1$ convolution maps.
(2) The part convolution maps are then processed with a distance transform, which we show is a generalization of max pooling.
(3) The root convolution map and the DT transformed part convolution maps are stacked into a single feature map with $P+1$ channels and then convolved with a sparse object geometry filter (see sidebar diagram and \secref{geo-filter}).
The output is a single-channel score map for the DPM component.
}
\figlabel{dpm-cnn-comp}
\end{figure*}

\subsubsection{Object geometry filters}
\seclabel{geo-filter}

The score of a DPM component $c$ at each root filter location $s$ is given 
by adding the root filter score at $s$ to the distance transformed part scores at ``anchor'' locations offset from $s$.
Each part $p$ has its own anchor offset that is specified by a 2D vector $v_p = (v_{px}, v_{py})$.

Computing component scores at all root locations can be rephrased as a convolution.
The idea is to stack the root filter score map together with the $P$ distance transformed part score maps to form a feature map with $P+1$ channels, and then convolve that feature map with a specially constructed filter that we call an ``object geometry'' filter.
This name comes from the fact that the filter combines a spatial configuration of parts into a whole.
The coefficients in the object geometry filter are all zero, except for a single coefficient set to one in each of the filter's $P+1$ channels.
The first channel corresponds to the root filter, and has a one in its upper-left corner (the root's anchor is defined as $v_0 = (0,0)$).
Filter channel $p$ (counting from zero), has a one at index $v_p$ (using matrix-style indexing, where indices grow down and to the right).
To clarify this description, an example object geometry filter for a DPM component with one root and one part is shown in the \figref{dpm-cnn-comp} sidebar.

DPM is usually thought of as a flat model, but making the object geometry filter explicit reveals that DPM actually has a second, implicit convolutional layer.
This insight shows that in principle one could train this filter discriminatively, rather than heuristically setting it to a sparse binary pattern.
We revisit this idea when discussing our experiments in \secref{experiments_prelim}.

\subsubsection{Combining mixture components with maxout}

Each single-component DPM-CNN produces a score map for each pyramid level.
Let $z_{sc}$ be the score for component $c$ at location $s$ in a pyramid level.
In the DPM formalism, components compete with each other at each location.
This competition is modeled as a max over component scores: $z_s = \max_c z_{sc}$, where the overall DPM score at $s$ is given by $z_s$.
In DPM-CNN, $z_{sc} = \bfw_c \cdot \bfx_s + b_c$, where $\bfw_c$ is component $c$'s object geometry filter (vectorized), $\bfx_s$ is the sub-array of root and part scores at $s$ (vectorized), and $b_c$ is the component's scalar bias.
\figref{dpm-cnn-multicomp} shows the full multi-component DPM-CNN including the max non-linearity that combines component scores.

It is interesting to note that the max non-linearity used in DPM is mathematically equivalent to the ``maxout'' unit recently described by Goodfellow \etal \cite{maxout}.
In the case of DPM, the maxout unit has a direct interpretation as a switch over the choice of model components.

\begin{figure*}[t!]
\centering
\includegraphics[scale=0.6]{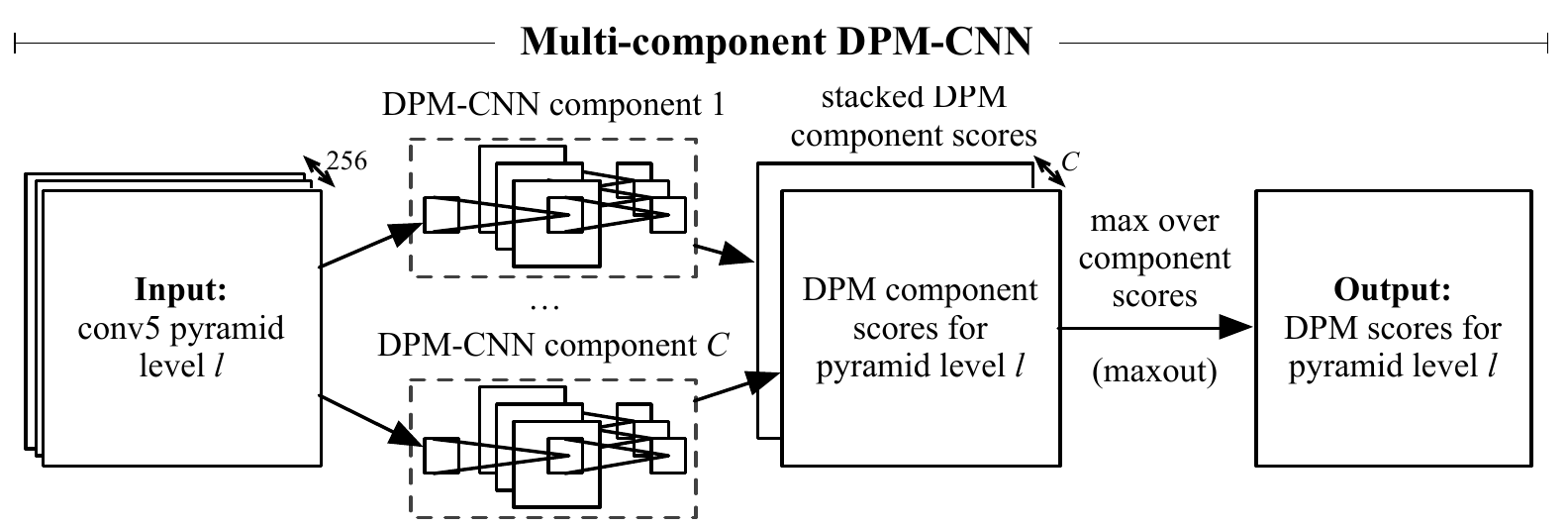}
\caption{\textbf{CNN equivalent to a multi-component DPM.} 
A multi-component DPM-CNN is composed of one DPM-CNN per component (\figref{dpm-cnn-comp}) and a maxout \cite{maxout} layer that takes a max over component DPM-CNN outputs at each location.
}
\figlabel{dpm-cnn-multicomp}
\end{figure*}


\subsubsection{Generalizations}

For clarity, we constructed a DPM-CNN for the case of mixtures of star models.
However, the construction is general and naturally extends to a variety of models such as object detection grammars \cite{girshick2011grammar} and recent DPM variants, such as \cite{yang2012articulated}.

The CNN construction for these more complex models is analogous to the DPM-CNN construction: take the exact inference algorithm used for detection and explicitly unroll it (this can be done given a fixed model instance); express the resulting network in terms of convolutional layers (for appearance and geometry filters), distance transform pooling layers, subsampling layers, and maxout layers.

We also note that our construction is limited to models for which exact inference is possible with a non-iterative algorithm.
Models with loopy graphical structures, such as Wang \etal's hierarchical poselets model \cite{wang2011learning}, require iterative, approximate inference algorithms that cannot be converted to equivalent fixed-depth CNNs.

\subsection{Related work}

Our work is most closely related to the deep pedestrian detection model of Ouyang and Wang \cite{ouyang13deep}.
Their CNN is structured like a DPM and includes a ``deformation layer'' that takes a distance penalized global max within a detection window.
Their work reveals some connections between single-component DPMs and CNNs, however since that is not the focus of their paper a clear mapping of generic DPMs to CNNs is not given.
We extend their work by: developing object geometry filters, showing that multi-component models are implemented with maxout, describing how DPM inference over a whole image is efficiently computed in a CNN by distance transform pooling, and using a more powerful CNN front-end to generate the feature pyramid.
Our DT-pooling layer differs from the ``deformation layer'' described in \cite{ouyang13deep} since it efficiently computes a distance transform over the entire input, rather than an independent global max for each window.

The idea of unrolling (or ``time unfolding'') an inference algorithm in order to construct a fixed-depth network was explored by Gregor and LeCun in application to sparse coding \cite{gregor2010learning}.
In sparse coding, inference algorithms are iterative and converge to a fixed point. 
Gregor and LeCun proposed to unroll an inference algorithm for a fixed number of iterations in order to define an approximator network.
In the case of DPM (and similar low tree-width models), the inference algorithm is exact and non-iterative, making it possible to unroll it into a fixed-depth network without any approximations.

Boureau \etal \cite{boureau2010theoretical} study average and max pooling from theoretical and experimental perspectives.
They discuss variants of pooling that parametrically transition from average to max.
Distance transform pooling, unlike the pooling functions in \cite{boureau2010theoretical}, is interesting because it has a learnable pooling region.
Jia \etal \cite{jia2012beyond} also address the problem of learning pooling regions by formulating it as a feature selection problem.

Our work is also related to several recent approaches to object detection using CNNs.
OverFeat \cite{overfeat} performs coarse sliding-window detection using a CNN.
At each rough location, OverFeat uses a regressor to predict the bounding box of a nearby object.
Another recent CNN-based object detector called DetectorNet \cite{DetectorNet} also performs detection on a coarse set of sliding-window locations.
At each location, DetectorNet predicts masks for the left, right, top, bottom, and whole object.
These predicted masks are then clustered into object hypotheses.
Currently, the most accurate object detection method for both ImageNet detection as well as PASCAL VOC is the Region-based CNN framework (R-CNN) \cite{rcnnTR,girshick2014rcnn}.
Unlike DetectorNet and OverFeat, R-CNN does not perform sliding-window detection; instead R-CNN begins by extracting a set of region proposals \cite{UijlingsIJCV2013} and classifies them with a linear SVM applied to CNN features.

\section{Implementation details}
\seclabel{impl}

We implemented our experiments by modifying the DPM voc-release5 code \cite{release5} and using Caffe \cite{Jia13caffe} for CNN computation.
We describe the most important implementation details in this section.
Source code for the complete system will be available.

\subsection{Feature pyramid construction}

Our DeepPyramid DPM implementation starts by processing an input image with a truncated variant of the SuperVision CNN \cite{krizhevsky2012imagenet}.
We use the publicly available network weights that are distributed with R-CNN \cite{girshick2014rcnn}, which allows for a direct comparison.
These weights were trained on the ILSVRC 2012 classification training dataset using Caffe (we do not use the detection fine-tuned weights).

Starting from this network, we made two structural modifications.
The first was to remove the last max pooling layer (\pool{5}) and all of the fully-connected layers (\fc{6}, \fc{7}, \fc{8}, and softmax).
The network's output is, therefore, the feature map computed by the fifth convolutional layer (\conv{5}), which has 256 feature channels.
The second modification is that before each convolutional or max pooling layer, with kernel size $k$, we zero-pad the layer's input with $\lfloor k/2 \rfloor$ zeros on all sides (top, bottom, left, and right).
This padding implements ``same'' convolution (or pooling), where the input and output maps have the same spatial extent.
With this padding, the mapping between image coordinates and CNN output coordinates is straightforward.
A ``pixel'' (or cell) at zero-based index $(x,y)$ in the CNN's \conv{5} feature map has a receptive field centered on pixel $(16x,16y)$ in the input image.
The \conv{5} features, therefore, have a stride of 16 pixels in the input image with highly overlapping receptive fields of size $163 \times 163$ pixels.
Our experimental results show that even though the receptive fields are very large, the features are localized well enough for sliding-window detectors to precisely localize objects without regression (as in OverFeat and DetectorNet).


For simplicity, we process the image pyramid with a naive implementation in which each image pyramid level is embedded in the upper-left corner of a large ($1713 \times 1713$ pixel) image.
For the first image pyramid level, the original image is resized such that its largest dimension is 1713 pixels.
For PASCAL VOC images, this results in up-sampling images by a factor of 3.4 on average.
The first \conv{5} pyramid level has 108 cells on its longest side.
We use a pyramid with 7 levels, where the scale factor between levels is $2^{-1/2}$ (the pyramid spans three octaves).
The entire \conv{5} pyramid has roughly 25k output cells (sliding-window locations).
For comparison, this is considerably more than the roughly 1,500 sliding-window locations used in OverFeat, but many fewer than the 250k commonly used in HOG feature pyramids.
Computing the \conv{5} feature pyramid as described above is fast, taking only 0.5 seconds on an NVIDIA Tesla K20c.


\subsection{Parameter learning}

DeepPyramid DPM is a single CNN that takes an image pyramid as input and outputs a pyramid of DPM detection scores. 
In principle, the entire network can be trained end-to-end using backpropagation by defining a loss function on the score pyramid.
In this work we opt for a simpler approach in which DeepPyramid DPM is trained stage-wise, in two stages.
The first stage trains the truncated SuperVision CNN.
For this we simply use the publicly available model distributed with R-CNN.
This model was trained for image classification on ILSVRC 2012 and was not fine-tuned for detection.
In the second stage, we keep the SuperVision CNN parameters fixed and train a DPM on top of \conv{5} feature pyramids using the standard latent SVM training algorithm used for training DPMs on HOG features.
In the future, we plan to experiment with training the entire system with backpropagation.

\subsection{DPM training and testing details}

Compared to training DPM with HOG features, we found it necessary to make some changes to the standard DPM training procedure.
First, we don't use left/right mirrored pairs of mixture components.
These components are easy to implement with HOG because model parameters can be explicitly ``flipped'' allowing them to be tied between mirrored components.
Second, R-CNN and DPM use different non-maximum suppression functions and we found that the one used in R-CNN, which is based on intersection-over-union (IoU) overlap, performs slightly better with \conv{5} features (it is worse for the baseline HOG-DPM).
Lastly, we found that it's very important to use poorly localized detections of ground-truth objects as negative examples.
As in R-CNN, we define negative examples as all detections that have a max IoU overlap with a ground-truth object of less than 0.3.
Using poorly localized positives as negative examples leads to substantially better results (\secref{ablation}) than just using negatives from negative images, which is the standard practice when training HOG-DPM.

\section{Experiments}
\seclabel{experiments}

\subsection{Results on PASCAL VOC 2007}
\seclabel{experiments_prelim}

Deformable part models have been tuned for use with HOG features over several years. 
A priori, it's unclear if the same structural choices that work well with HOG (\eg, the number of mixture components, number of parts, multi-resolution modeling, \etc) will also work well with very different features.
We conducted several preliminary experiments on the PASCAL VOC 2007 dataset \cite{PASCAL07} in order to find a DPM structure suited to \conv{5} features.


In \tableref{prelim}, rows 1-3, we show the effect of adding parts to a three component DPM (three was selected through cross-validation).
As in HOG-based DPMs, the dimensions of root filters vary across categories and components, influenced by the aspect ratio distribution for each class.
Our root filter sizes typically range from $4 \times 12$ to $12 \times 4$ feature cells.
We start with a ``root-only'' model (\ie, no parts) and then show results after adding 4 or 8 parts (of size $3 \times 3$) to each component.
With 4 parts, mAP increases by 0.9 percentage points, with an improvement in 16 out of 20 classes.
The effect size is small, but appears to be statistically significant ($p = 0.016$) as judged by a paired-sample permutation test, a standard analysis tool used in the information retrieval community \cite{IRsignificance}.

One significant difference between HOG and \conv{5} features is that HOG describes scale-invariant local image statistics (intensity gradients), while \conv{5} features describe large ($163 \times 163$ pixel) image patches.
The top two rows of \figref{pyramid} illustrate this point.
Each row shows a feature pyramid for an image of a face.
The first is HOG and the second is the ``face channel'' of \conv{5}.
In the HOG representation, the person's face appears at all scales and one can imagine that for each level in the pyramid, it would be possible to define an appropriately sized template that would fire on the face.
The \conv{5} face channel is quite different.
It only shows strong activity when the face is observed in a narrow range of scales.
In the first several levels of the pyramid, the face feature responses are nearly all zero (black).
The feature peaks in level 6 when the face is at the optimal scale.

Based on the previous experiments with parts and the feature visualizations, we decided to explore another hypothesis: that the convolution filters in \conv{5} already act as a set of shared ``parts'' on top of the \conv{4} features.
This perspective suggests that one can spread the \conv{5} features to introduce some local deformation invariance and then learn a root-only DPM to selectively combine them.
This hypothesis is also supported by the features visualized in \figref{pyramid}.
The heat maps are characteristic of part responses (cat head, person face, upper-left quarter of a circle) in that they select specific visual structures.

We implemented this idea by applying a $3 \times 3$ max filter to \conv{5} and then training a root-only DPM with three components.
The max filter is run with a stride of one to prevent subsampling the feature map, which would increase the sliding-window stride to a multiple of 16.
We refer to the max-filtered \conv{5} features as ``\maxp{5}''.
Note that this model does not have any explicit DPM parts and that the root filters can be thought of as a \emph{learned} object geometry filters that combine the \conv{5} ``parts''.
This approach (\tableref{prelim} row 4) outperforms the DPM variants that operate directly on \conv{5} in terms of mAP as well as training and testing speed (since each model only has three filters).

\begin{figure*}[t!]
\begin{centering}
\includegraphics[width=1\linewidth]{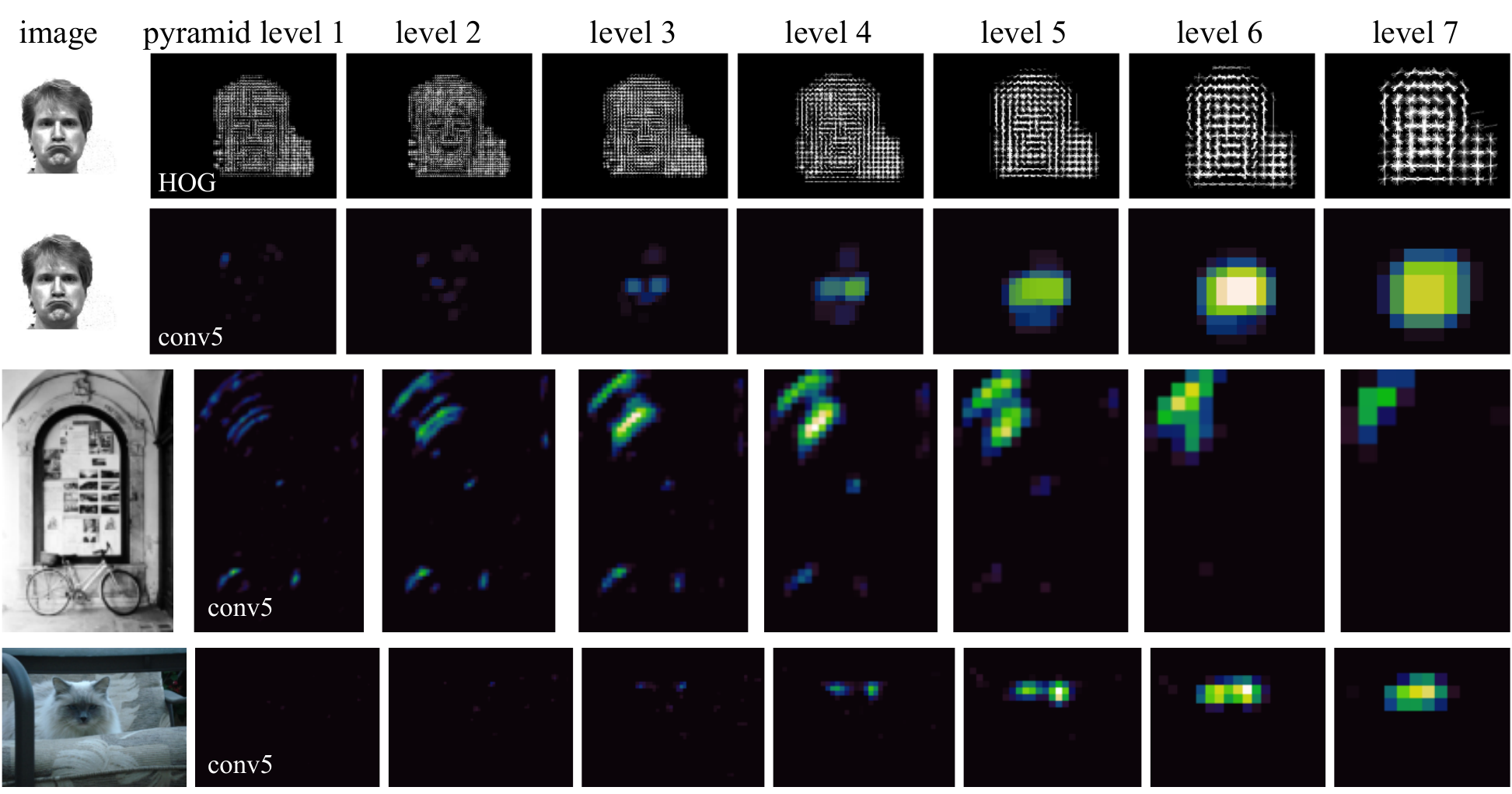}
\end{centering}
\vspace{-1em}
\caption{\textbf{HOG versus \conv{5} feature pyramids.} 
In contrast to HOG features, \conv{5} features are more part-like and scale selective.
Each \conv{5} pyramid shows 1 of 256 feature channels.
See \secref{experiments_prelim} for details.
}
\figlabel{pyramid}
\end{figure*}

\subsubsection{Comparison with other methods}

We compare our approach to several other methods on the VOC 2007 dataset.
The first notable comparison is to DPM on HOG features.
We report results using the standard 6 component, 8 part configuration, which achieves a mAP of 33.7\%.
We also computed another HOG-DPM baseline using 6 components, but without parts.
Removing parts decreases HOG-DPM performance to 25.2\%.
The \maxp{5} variant of DeepPyramid DP, which uses \conv{5} implicitly as a set of shared parts, has a mAP of 45.2\%.

We also compare our method to the recently proposed R-CNN \cite{girshick2014rcnn}.
The directly comparable version of R-CNN uses \pool{5} features and no fine-tuning (\pool{5} is the same as \maxp{5}, but with a stride of two instead of one).
This comparison isolates differences to the use of a sliding-window method versus classifying warped selective search \cite{UijlingsIJCV2013} windows.
We can see that for some classes where we expect segmentation to succeed, such as aeroplane and cat, R-CNN strongly outperforms DeepPyramid DPM.
For classes where we expect segmentation might fail, such as bottle, chair, and person, DeepPyramid DPM strongly outperforms R-CNN.
Performance is similar for most of the other categories, with DeepPyramid DPM edging out R-CNN \pool{5} in terms of mAP.
Of course, this represents the weakest variant of R-CNN, with significant improvements coming from fine-tuning for detection and incorporating a bounding-box (BB) regression stage.

We have shown that DeepPyramid DPM is competitive with R-CNN \pool{5} without fine-tuning.
The R-CNN results suggest that most of the gains from fine-tuning come in through the non-linear classifier (implemented via layers \fc{6} and \fc{7}) applied to \pool{5} features.
This suggests that it might be possible to achieve similar levels of performance with DeepPyramid DPM through the use of a more powerful non-linear classifier.

\begin{table*}[t!]
\centering
\caption{\textbf{Detection average precision (\%) on VOC 2007 test.}
Column $C$ shows the number of components and column $P$ shows the number of parts per component.
Our method is DP-DPM (DeepPyramid DPM).
}
\renewcommand{\arraystretch}{1.3}
\renewcommand{\tabcolsep}{1.1mm}
\resizebox{\linewidth}{!}{
\begin{tabular}{@{}l|c|c|r*{19}{c}|c@{}}
method & $C$ & $P$ & aero      & bike      & bird      & boat      & botl     & bus        & car        & cat        & chair      & cow        & table      & dog        & horse      & mbike      & pers     & plant      & sheep      & sofa       & train      & tv         & mAP       \\
\hline
DP-DPM \conv{5}  &  3  &  0  &  41.2  &  64.1  &  30.5  &  23.9  &  35.6  &  51.8  &  54.5  &  37.2  &  25.8  &  46.1  &  38.8  &  39.1  &  58.5  &  54.8  &  51.6  &  25.8  &  \bf{48.3}  &  33.1  &  49.1  &  56.1  &  43.3 \\
DP-DPM \conv{5}  &  3  &  4  &  43.1  &  64.2  &  31.3  &  25.0  &  \bf{37.6}  &  55.8  &  55.7  &  37.8  &  27.3  &  46.0  &  35.5  &  39.0  &  58.2  &  57.0  &  53.0  &  \bf{26.6}  &  47.6  &  35.3  &  50.6  &  56.6  &  44.2 \\
DP-DPM \conv{5}  &  3  &  8  &  42.3  &  65.1  &  32.2  &  24.4  &  36.7  &  \bf{56.8}  &  55.7  &  38.0  &  \bf{28.2}  &  47.3  &  37.1  &  39.2  &  \bf{61.0}  &  56.4  &  52.2  &  \bf{26.6}  &  47.0  &  35.0  &  51.2  &  56.1  &  44.4 \\
DP-DPM \maxp{5}  &  3  &  0  &  44.6  &  \bf{65.3}  &  32.7  &  24.7  &  35.1  &  54.3  &  56.5  &  40.4  &  26.3  &  \bf{49.4}  &  43.2  &  \bf{41.0}  &  \bf{61.0}  &  55.7  &  \bf{53.7}  &  25.5  &  47.0  &  \bf{39.8}  &  47.9  &  \bf{59.2}  &  \bf{45.2} \\
\hline
HOG-DPM &  6  &  0  &  23.8  &  51.3  &  \phz5.1  &  11.5  &  19.2  &  41.3  &  46.3  &  \phz8.5  &  15.8  &  20.8  &  \phz8.6  &  10.4  &  43.9  &  37.6  &  31.9  &  11.9  &  18.1  &  25.7  &  36.5  &  35.4  &  25.2 \\
HOG-DPM \cite{release5}  &  6  &  8  &  33.2  &  60.3  &  10.2  &  16.1  &  27.3  &  54.3  &  58.2  &  23.0  &  20.0  &  24.1  &  26.7  &  12.7  &  58.1  &  48.2  &  43.2  &  12.0  &  21.1  &  36.1  &  46.0  &  43.5  &  33.7 \\
HSC-DPM \cite{HSC}  &  6  &  8  &  32.2  &  58.3  &  11.5  &  16.3  &  30.6  &  49.9  &  54.8  &  23.5  &  21.5  &  27.7  &  34.0  &  13.7  &  58.1  &  51.6  &  39.9  &  12.4  &  23.5  &  34.4  &  47.4  &  45.2  &  34.3 \\
DetectorNet \cite{DetectorNet}  &  n/a  &  n/a  &  29.2  &  35.2  &  19.4  &  16.7  &  \phz3.7  &  53.2  &  50.2  &  27.2  &  10.2  &  34.8  &  30.2  &  28.2  &  46.6  &  41.7  &  26.2  &  10.3  &  32.8  &  26.8  &  39.8  &  47.0  &  30.5 \\
R-CNN \cite{girshick2014rcnn} \pool{5} &  n/a  &  n/a  &  \bf{51.8}  &  60.2  &  \bf{36.4}  &  \bf{27.8}  &  23.2  &  52.8  &  \bf{60.6}  &  \bf{49.2}  &  18.3  &  47.8  &  \bf{44.3}  &  40.8  &  56.6  &  \bf{58.7}  &  42.4  &  23.4  &  46.1  &  36.7  &  \bf{51.3}  &  55.7  &  44.2 \\
\hline
\multicolumn{24}{@{}l@{}}{fine-tuned variants of R-CNN} \\
\hline
R-CNN FT \pool{5} & n/a & n/a  &  58.2  &  63.3  &  37.9  &  27.6  &  26.1  &  54.1  &  66.9  &  51.4  &  26.7  &  55.5  &  43.4  &  43.1  &  57.7  &  59.0  &  45.8  &  28.1  &  50.8  &  40.6  &  53.1  &  56.4  &  47.3 \\
R-CNN FT \fc{7}  & n/a & n/a &  64.2  &  69.7  &  50.0  &  41.9  &  32.0  &  62.6  &  71.0  &  60.7  &  32.7  &  58.5  &  46.5  &  56.1  &  60.6  &  66.8  &  54.2  &  31.5  &  52.8  &  48.9  &  57.9  &  64.7  &  54.2 \\
R-CNN FT \fc{7} BB  & n/a & n/a &  \bf{68.1}  &  \bf{72.8}  &  \bf{56.8}  &  \bf{43.0}  &  \bf{36.8}  &  \bf{66.3}  &  \bf{74.2}  &  \bf{67.6}  &  \bf{34.4}  &  \bf{63.5}  &  \bf{54.5}  &  \bf{61.2}  &  \bf{69.1}  &  \bf{68.6}  &  \bf{58.7}  &  \bf{33.4}  &  \bf{62.9}  &  \bf{51.1}  &  \bf{62.5}  &  \bf{64.8}  &  \bf{58.5} \\
\end{tabular}
}
\tablelabel{prelim}
\end{table*}

\subsubsection{Ablation study}
\seclabel{ablation}

To understand the effects of some of our design choices, we report mAP performance on VOC 2007 test using a few ablations of the DP-DPM \maxp{5} model.
First, we look at mAP versus the number of mixture components.
Mean AP with \{1, 2, 3\} components is \{39.9\%, 45.1\%, 45.2\%\}.
For most classes, performance improves when going from 1 to 2 or 1 to 3 components because the variety of templates allows for more recall.
We also looked at the effect of training with negative examples that come only from negative images (\ie, not using mislocalized positives as negative examples).
Using negatives only from negative images decreases mAP by 6.3 percentage points to 38.8\%.
Using standard DPM non-maximum suppression decreases mAP by 1.3 percentage points.

\subsection{Results on PASCAL VOC 2010-2012}

We used the VOC 2007 dataset for model and hyperparameter selection, and now we report results on VOC 2010-2012 obtained using the official evaluation server.
\tableref{voc2010} compares DeepPyramid DPM with a variety of methods on VOC 2010.
DeepPyramid DPM outperforms all recent methods other than the fine-tuned versions of R-CNN.
Performance against HOG-DPM is especially strong.
When comparing to R-CNN FT \fc{7}, without bounding-box regression (BB), DeepPyramid DPM manages better performance in two classes: bottle and person.
This likely speaks to the weakness in the region proposals for those classes.
The VOC 2011 and 2012 sets are the same and performance is similar to 2010, with a mAP of 41.6\%.

\begin{table*}[t!]
\centering
\caption{\textbf{Detection average precision (\%) on VOC 2010 test.}
}
\renewcommand{\arraystretch}{1.3}
\renewcommand{\tabcolsep}{1.1mm}
\resizebox{\linewidth}{!}{
\begin{tabular}{@{}l|r*{19}{c}|c@{}}
method        & aero      & bike      & bird      & boat      & botl     & bus        & car        & cat        & chair      & cow        & table      & dog        & horse      & mbike      & pers     & plant      & sheep      & sofa       & train      & tv         & mAP       \\
\hline
DP-DPM \maxp{5} & 61.0& 55.7& 36.5& 20.7& 33.2& 52.5& 46.1& 48.0& 22.1& 35.0& 32.3& 45.7& 50.2& 59.2& 55.8& 18.7& 49.1& 28.8& 40.6& 48.1& 42.0 \\
\hline
HOG-DPM \cite{release5}   &  49.2  &  53.8  &  13.1  &  15.3  &  35.5  &  53.4  &  49.7  &  27.0  &  17.2  &  28.8  &  14.7  &  17.8  &  46.4  &  51.2  &  47.7  &  10.8  &  34.2  &  20.7  &  43.8  &  38.3  &  33.4 \\
UVA \cite{UijlingsIJCV2013}  &  56.2  &  42.4  &  15.3  &  12.6  &  21.8  &  49.3  &  36.8  &  46.1  &  12.9  &  32.1  &  30.0  &  36.5  &  43.5  &  52.9  &  32.9  &  15.3  &  41.1  &  31.8  &  47.0  &  44.8  &  35.1 \\
Regionlets \cite{regionlets}  &  65.0  &  48.9  &  25.9  &  24.6  &  24.5  &  56.1  &  54.5  &  51.2  &  17.0  &  28.9  &  30.2  &  35.8  &  40.2  &  55.7  &  43.5  &  14.3  &  43.9  &  32.6  &  54.0  &  45.9  &  39.7 \\
SegDPM \cite{fidler2013bottom}  &  61.4  &  53.4  &  25.6  &  25.2  &  35.5  &  51.7  &  50.6  &  50.8  &  19.3  &  33.8  &  26.8  &  40.4  &  48.3  &  54.4  &  47.1  &  14.8  &  38.7  &  35.0  &  52.8  &  43.1  &  40.4 \\
R-CNN FT \fc{7} \cite{girshick2014rcnn} &  67.1  &  64.1  &  46.7  &  32.0  &  30.5  &  56.4  &  57.2  &  65.9  &  27.0  &  47.3  &  40.9  &  66.6  &  57.8  &  65.9  &  53.6  &  26.7  &  56.5  &  38.1  &  52.8  &  50.2  &  50.2 \\
R-CNN FT \fc{7} BB  &  \bf{71.8}  &  \bf{65.8}  &  \bf{53.0}  &  \bf{36.8}  &  \bf{35.9}  &  \bf{59.7}  &  \bf{60.0}  &  \bf{69.9}  &  \bf{27.9}  &  \bf{50.6}  &  \bf{41.4}  &  \bf{70.0}  &  \bf{62.0}  &  \bf{69.0}  &  \bf{58.1}  &  \bf{29.5}  &  \bf{59.4}  &  \bf{39.3}  &  \bf{61.2}  &  \bf{52.4}  &  \bf{53.7} \\
\end{tabular}
}
\tablelabel{voc2010}
\end{table*}

\section{Conclusion}
We have presented a novel synthesis of deformable part models and convolutional neural networks.
In particular, we demonstrated that any DPM can be expressed as an equivalent CNN by using distance transform pooling layers and maxout units.
Distance transform pooling generalizes max pooling and relates the idea of deformable parts to max pooling.
We also showed that a DPM-CNN can run on top a feature pyramid constructed by another CNN.
The resulting model---which we call \emph{DeepPyramid DPM}---is a single CNN that performs multi-scale object detection by mapping an image pyramid to a detection score pyramid.
Our theoretical and experimental contributions bring new life to DPM and show the potential for replacing HOG templates in a wide range of visual recognition systems.

\paragraph{Acknowledgments}
The GPUs used in this research were generously donated by the NVIDIA Corporation.

{\footnotesize
\bibliographystyle{ieee}
\bibliography{main}
}

\end{document}